\documentclass[runningheads]{llncs}
\usepackage[T1]{fontenc}
\usepackage[misc,geometry]{ifsym}

\usepackage{graphicx}
\usepackage{amsmath}
\usepackage{amsfonts}
\usepackage{amssymb}
\usepackage{array}
\usepackage{multirow}
\usepackage{fancyhdr}
\begin{document}

\title{From Concrete to Abstract: A Multimodal Generative Approach to Abstract Concept Learning}
%
%
\author{Haodong Xie\textsuperscript{(\Letter)} \and
Rahul Singh Maharjan\and
Federico Tavella\and
Angelo Cangelosi}
%
%

  \institute{Manchester Centre for Robotics and AI, University of Manchester, Manchester, United Kingdom\\
  \email{\{haodong.xie, rahulsingh.maharjan,  federico.tavella, angelo.cangelosi\}@manchester.ac.uk}}

%
\maketitle              

\thispagestyle{fancy}
\begin{abstract}
Understanding and manipulating concrete and abstract concepts is fundamental to human intelligence. Yet, they remain challenging for artificial agents. This paper introduces a multimodal generative approach to high order abstract concept learning, which integrates visual and categorical linguistic information from concrete ones. Our model initially grounds subordinate level concrete concepts, combines them to form basic level concepts, and finally abstracts to superordinate level concepts via the grounding of basic-level concepts. We evaluate the model language learning ability through language-to-visual and visual-to-language tests with high order abstract concepts. Experimental results demonstrate the proficiency of the model in both language understanding and language naming tasks.

\end{abstract}
%
%
\section{Introduction}

What would a drawing of an Artificial Intelligence (AI) agent look like if asked to illustrate an animal? 
Among the most frequently used words by humans, only $28\%$ are concrete, with the majority being abstract concepts \cite{Schwanenflugel}. \textit{Concrete concepts} have single, bounded, and identifiable referents that can be perceived through our senses, whereas  \textit{abstract concepts} lack such concrete and direct-sensory referents, making them hard to identify \cite{Borghi 2017}. This distinction makes abstract concepts more challenging to understand, process, acquire and remember compared to concrete ones \cite{Bates}. Despite being easily distinguishable from each other \cite{Katja}, abstract and concrete concepts are not strictly separate as a dichotomy but exist along a continuum \cite{Barsalou}. The level of concreteness increases from highly abstract concepts such as "freedom" to highly concrete concepts like "Border Collie". The continuum perspective suggests that learning higher-order, more abstract words can be facilitated by extending the strategies and models used for grounding concrete words. Following this idea, to develop models capable of understanding both abstract and concrete concepts, the learning process starts with understanding the concrete concepts. Once a sufficient number of concrete words are learned, concepts with a higher level of abstraction can be acquired by combining concrete concepts. However, few studies have explored this extension in the context of artificial intelligence \cite{Di Nuovo}.
On the contrary, abstract concepts are crucial, as most of the human's daily language is abstract \cite{Wiemer-Hastings}. The ability to understand and manipulate abstract concepts is a fundamental trait of human intelligence; however, this ability is currently lacking in artificial agents \cite{Di Nuovo}.

In this work, we aim to explore the development of higher-order concepts with visual and categorical linguistic information through the implementation of a multimodal generative model. Our method employs three levels of conceptual categorisation: subordinate, basic, and superordinate. Concepts at the subordinate level are specific, such as ``goldfish" within ``fish". These subordinate level concepts together form the basic level concepts, like ``fish". Then the superordinate level, such as "animal", encompasses a range of basic level concepts\cite{Esbrí}. The acquisition of abstract concepts is achieved through the learning of concrete concepts from the subordinate level. Then, the basic level concepts (more frequently used) are learned via the combination of concrete concepts. Finally, it progresses to grounding the superordinate level concepts (more abstract) through the combinatorial power of language with the basic level concepts. The level of abstraction increases from the subordinate level to the basic level, and then to the superordinate level concepts \cite{Rosch}. 
The multimodal generative model we propose in this work learns abstract concepts starting from concrete concepts. After the model has learned enough concrete concepts, it can understand more abstract concepts. This approach provides a way for AI agents to learn abstract concepts through the understanding of concrete concepts. At the same time, as vision is human's most important sense \cite{Hutmacher}, our model learns the concepts with both visual and linguistic information. To the best of our knowledge, this is the first study to explore the development of high order concepts under the visual-linguistic multimodal setting.

\begin{figure}
\includegraphics[width=\textwidth]{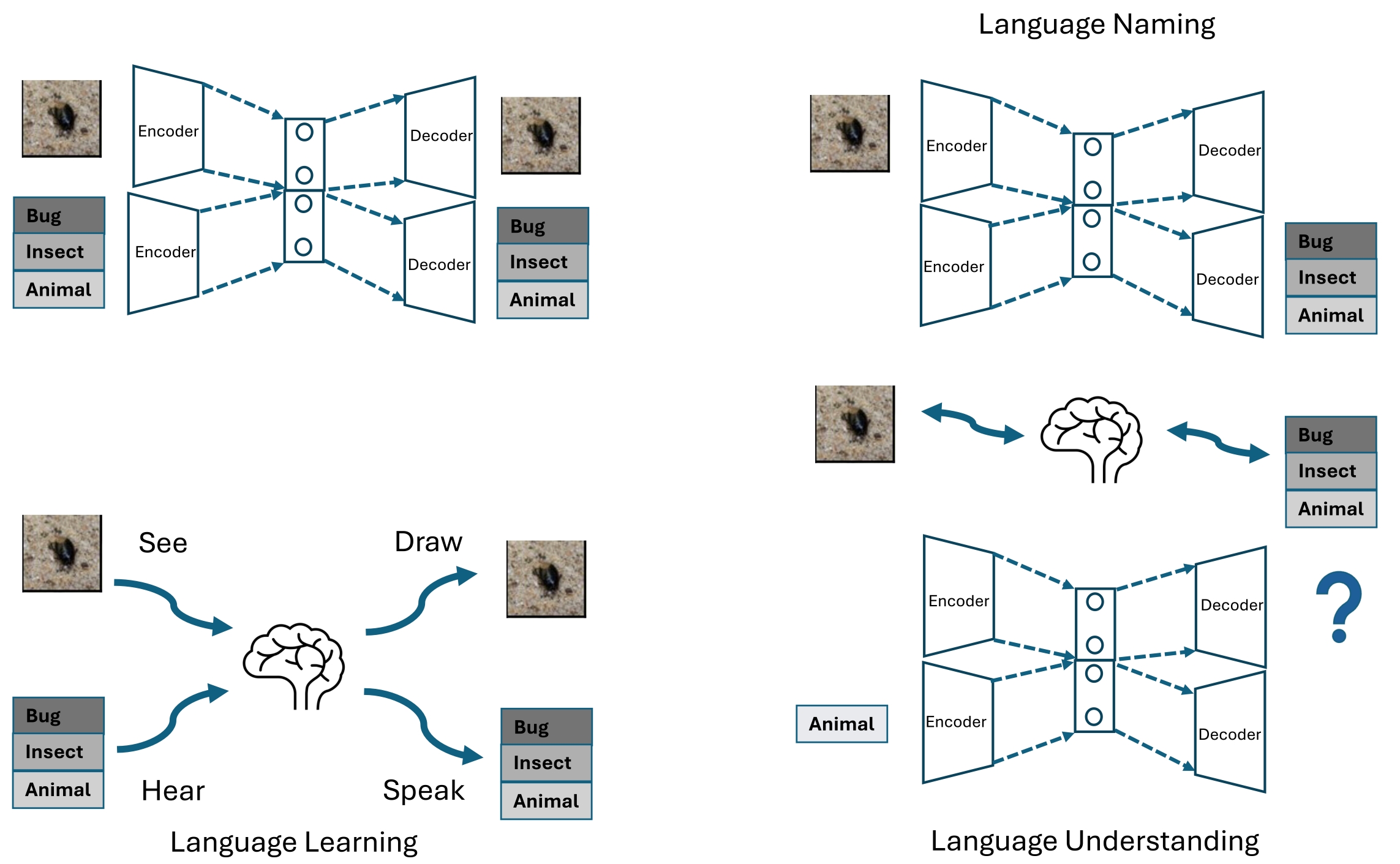}
\caption{The Joint Modalities Learning and Cross Modality Test. During training, the model learns the concepts jointly with both visual and linguistic information. During testing, when one modality is missing, the model generates the missing modality based on the input modality.} 
\label{fig1}
\end{figure}

\section{Related Work}

\subsection{Abstract Concepts Learning with Computational Models}

Most existing grounded language models focus on learning words associated with concrete concepts, with limited research dedicated to computational models for abstract language learning.
Stramandinoli et al.\cite{Stramandinoli,Stramandinoli 2} use neural networks to learn concrete motor concepts through sensorimotor experience, and  ground higher-order action concepts via the combination of concrete motor concepts. Their work suggests that a hierarchical organization of concepts could facilitate the acquisition of abstract words. Di Nuovo et al. \cite{Di Nuovo A,Ruciński} explore how embodied strategies, such as finger use, enhance an agent's understanding of numerical concepts.  For emotion concepts learning, Prescott et al.\cite{Prescott} proposed a Gaussian process latent variable model to create a multimodal memory system for the iCub robot. This system enables the retrieval of past events along with their associated emotions. The models described so far \cite{Stramandinoli,Stramandinoli 2,Di Nuovo A,Ruciński,Prescott} were able to learn different types of abstract concepts. 
However, the acquisition of more abstract concepts remains challenging for computational models. The approach we apply in this work to ground the higher-order concepts through the combination of concrete concepts. It provides a way to include more concrete and abstract concepts for the acquisition of abstract words. After the model has learned enough concrete concepts, it can then understand more abstract concepts. This can be a potential way for computational models to learn more abstract concepts.

\subsection{Multimodal Generative Model for Language-Visual Generation}

Numerous studies ventured to derive shared representations from diverse modalities and enable cross-modal generation using these representations. Some works \cite{Wu,Vedantam,Suzuki M,Pandey G,Jo D U,Yadav R,Shi Y,Shi} are based on the Variational Autoencoder (VAE) \cite{Kingma D P}, while others \cite{Huang,Zhang,Liu M,Zhang H,Xia W,Xu T} are based on the Generative Adversarial Network (GAN) \cite{Goodfellow I}. VAEs have become mainstream in multimodal deep generative models due to their ability to handle high-dimensional inputs, incorporate general-purpose priors for effective representation learning, and deal with the heterogeneity of multimodal data. In contrast, GANs struggle to introduce general-purpose priors as flexibly as VAEs in the multimodal setting, and sometimes suffer from instability during training \cite{Suzuki}.
To learn the joint distribution of the language and visual modalities, some researchers \cite{Wu,Vedantam,Huang,Zhang} leverage a product-of-experts (PoE) inference network. POE combines the outputs of multiple experts by multiplying their individual probability distributions, yielding a joint distribution with greater confidence. An alternative approach is the mixture-of-experts network (MoE), which is used in MMVAE \cite{Shi Y,Shi}. The MoE network combines the outputs of experts by summing their weighted contributions, resulting in a joint distribution that is computed with the most relevant experts. Compared to PoE, MoE's additive nature facilitates the optimization of the individual experts and mitigates potentially overconfident experts \cite{Sutter T}. 
In vision and language multimodal models that work with complex images, the visual modality often contains significantly more information than the language modality due to the large size and multi dimensional nature of images. This creates a complexity imbalance between the different modalities. Consequently, overconfident experts may occur in a PoE network, causing the model to focus more on the image modality than on the language modality when factoring the joint posterior distribution. For this reason, in this work, we choose a MoE network as the approach to integrate the language and visual modalities in our model.

\section{Methodology}

In this paper, we propose a generative model for the development of superordinate and basic level concepts grounded on subordinate level concepts. Our model is based on a Multimodal Mixture-of-Experts Variational Autoencoders (MMVAE) \cite{Shi} , enabling concept learning with both linguistic representation and visual information. The model also possesses cross-modal generation capabilities to provide post-training language-to-vision and vision-to language-tests. The learning and testing process of the model is shown in Figure 1. The model is trained with information from both modalities and is evaluated based on the accuracy of generating the missing modality from the existing modality.

\subsection{Variational Autoencoder for Visual Modality}
The training of VAE\cite{Kingma D P} aims to maximize the marginal likelihood \( p_\Theta(x) \), which is the probability of the observed data, $x$ under the model parameters \(\Theta\). As the true joint posterior distribution $p_\Theta(z \mid x)$ is not directly accessible, computing the marginal likelihood requires integrating over all possible values of the latent variable $z$. Consequently, the marginal likelihood becomes intractable. In this case, the evidence lower bound (ELBO) is instead optimized, and a variational posterior $q_\Phi(z \mid x)$ is used to approximate the true posterior $p_\Theta(z \mid x)$. ELBO is formed by an expected log-likelihood of the data and Kullback-Leibler (KL) divergence between the variational posterior $q_\Phi(z \mid x)$ and the prior distribution $p_{\Theta}(z)$

\begin{equation}
\text{ELBO}(x) = \mathbb{E}_{z \sim q_{\Phi}(z \mid x)}[ \log p_{\Theta}(x \mid z)] - \text{KL}(q_{\Phi}(z \mid x) \parallel p_{\Theta}(z))
\end{equation}

We use VAE to learn the distribution of visual information. For model input, we select the classes of images associated with subordinate level concepts in the language modality information from ImageNet \cite{Olga Russakovsky} . Following Shi et al. \cite{Shi}, we also perform feature-to-feature regeneration for the reconstruction of visual modality information to mitigate problems with producing blurry images in complex datasets. Instead of generating directly in the image space, we generate within the feature space of a pre-trained ResNet-101. The ResNet-101 produces the feature map of the image. The image VAE takes this 2048-dimension feature map as input. 
The image encoder is composed of three linear layers (with dimensions 256, 512, and 1024), followed by two additional linear layers, each with 128 dimensions, to project the output to the latent space. The decoder consists of four linear layers (with dimensions 256, 512, 1024, and 2048) to regenerate the feature maps. We utilize a nearest-neighbor lookup method in the feature space, employing Euclidean distance to match the reconstructed features with the closest existing features.

\subsection{Variational Autoencoder for Language Modality}

Due to the complexity imbalance between the language and visual modalities (i.e., a $224\times224$ image for the visual modality versus 2 labels for the language modality), when  the visual modality is missing, the information provided by the language modality is insufficient to reconstruct the visual modality. This makes reconstructing the visual modality from the language modality in cross-modality generation challenging. Unlike models\cite{Shi Y,Shi,Huang,Zhang,Xia W,Xu T} that generate caption descriptions of images, our model generates labels. The limited vocabulary size of words and the insufficient relationships between words make it inadequate for training a well-performing embedding layer. To address this, we incorporate a pre-trained BERT \cite{Devlin} network before the language VAE. BERT generates the embeddings for the concepts, and its parameters are frozen during the autoencoder's training process. 
The labels are first embedded by BERT, and then these embeddings serve as the input for the VAE. On the decoder side, the resampled joint distribution is taken as input and decoded into reconstructed embedding vectors. The language encoders consist of four convolutional layers with a $1\times 4$ kernel size, followed by an additional convolutional layers with a $3\times 3$ kernel size to project the output to the specified latent space dimension. After embedding the words using the pre-trained BERT model, the embeddings are encoded by the language encoder. The language decoders are composed of five convolutional layers and one linear layer to transform the vector back into words.
As shown in Table 1, the language modality inputs are three levels of hierarchical labels pair to pair to the visual modality information. The level of abstraction increases from the subordinate level concrete concepts to the basic level concept, then to the superordinate abstract concepts.

\begin{table}
\caption{The language modality consists of three hierarchical levels of concepts: the superordinate level, the basic level, and the subordinate level. Each superordinate level category contains five basic level categories, and each basic level category is formed by three subordinate level categories.}\label{tab1}
\begin{tabular}{|>{\centering\arraybackslash}m{4cm}|>{\centering\arraybackslash}m{4cm}|>{\centering\arraybackslash}m{4cm}|}
\hline
  {\large Superordinate Level} &    {\large Basic Level} &   {\large Subordinate Level}\\
\hline
\multirow{9}{*}{Animal} &  \multirow{3}{*}{Fish}  & Goldfish\\
       && Shark\\
       && Tuna\\\cline{2-3}
       &\multirow{3}{*}{Horse} & Mule\\
       && Pony\\
       && Zebra\\\cline{2-3}
       & \multirow{3}{*}{Squirrel}  & Chipmunk\\
       && Gopher\\
       && Marmot\\\cline{2-3}
       &  \multirow{3}{*}{Bird}& Chicken\\
       && Parrot\\
       && Swallow\\\cline{2-3}
       &  \multirow{3}{*}{Insect}& Bug\\
       && Butterfly\\
       && Fly\\
\hline
\end{tabular}
\end{table}
\subsection{Multimodal Variational Autoencoder}

To learn a joint distribution which represent the information from every modalities, a MoE network is used. MoE computes the variational posterior \( q_\Phi(z \mid x_{1:M}) \)  as a weighted sum of the individual posteriors \( q_{\phi_m}(z \mid x_m) \) for each modality. In our work, we assume that after being embedded by BERT, the information on language modality will have a similar level of complexity as information on the visual modality, so each modality shares the same weight value for the joint distribution computation.

\begin{equation}
q_\Phi(z \mid x_{1:M}) = \sum_{m=1}^{M} \frac{1}{M} \cdot q_{\phi_m}(z \mid x_m)
\end{equation}

In the multimodal setting with MoE, the ELBO loss includes the reconstruction term $\mathbb{E}_{z_m \sim q_{\phi_m}(z \mid x_m)} \left[ \log p_\Theta(x_m \mid z_m) \right]$ that represents the accuracy with which each modality can reconstruct the information $x_m$ from sample distribution $z$; Kullback-Leibler (KL) divergence between the variational posterior $q_\Phi(z \mid x_m)$ and the $p_\Theta(z)$.

\begin{equation}
\begin{aligned}
\text{ELBO}(x_{1:M}) = \frac{1}{M} \sum_{m=1}^{M} [ \mathbb{E}_{z_m \sim q_{\phi_m}(z \mid x_m)} [ \log p_\Theta(x_m \mid z_m) ]\\ - \text{KL}(q_{\phi_m}(z \mid x_m) \parallel p_\Theta(z)) ]
\end{aligned}
\end{equation}

The language and visual information are paired and input into the network. An encoder processes each modality's information into a latent space vector. Then, single-modal distributions are integrated into a multi-modal probability distribution using a MoE. Decoders for each modality can then generate information from the joint distribution. After training, the model can not only reconstruct the observed modality but also generate the missing modality based on the observed information from the other modality. The architecture of the proposed model is shown as Fig.2. For the model training, we choose Adam as optimizer with a learning rate of 0.001 and a 128 dimension latent space to represent the multimodal information.

\begin{figure}
\includegraphics[width=\textwidth]{
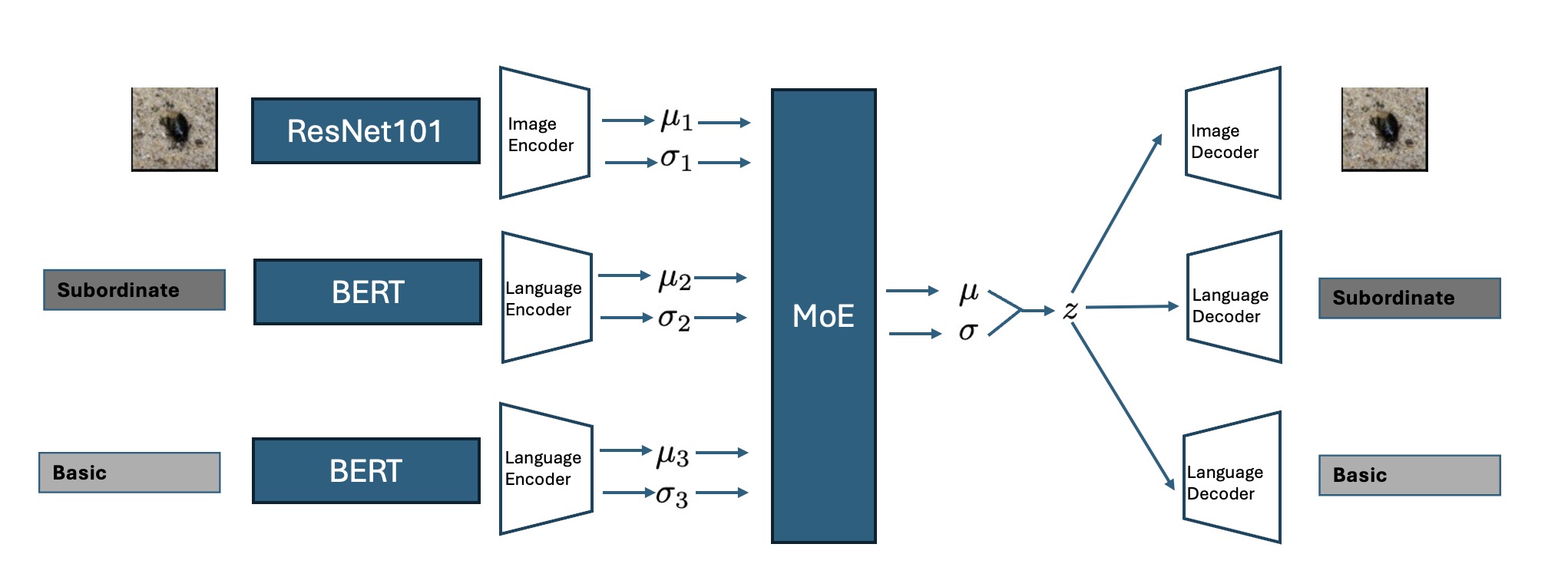}
\caption{The overall structure of our model: similar to MVAE19 proposed by Wu et al.\cite{Wu}, which treats each attribute as its own modality, we consider each level of concepts as its own modality and apply two separate language VAEs for the label input. A joint distribution representing the information from these four modalities is generated by the MoE network. The decoders sample from the latent space and then regenerate the image and the two levels of concepts.} \label{fig1}
\end{figure}

\section{Experiment}

We evaluate the performance of our model with two tests to assess how well it has acquired knowledge of the concepts: a language-to-vision test and a vision-to-language test.
To compute the accuracy of the generated results from the language-to-vision test, we train a classifier composed of a pre-trained VGG19 model \cite{Simonyan} and five trainable layers. The classifier is trained using the subset we selected from ImageNet and the paired hierarchical concepts. It takes images as input and predicts the hierarchical labels. In the language understanding test, the model generates the missing visual modality information as an image. The trained classifier receives the image as input and predicts the labels. We compute the cross-modality accuracy based on the comparison between the input label of our model and the predicted label from the classifier. 
In the language naming test, our model takes the image as input and generates labels. These generated labels are compared with the true labels of the image to assess accuracy. Also, we use the Clip score\cite{Hessel}  to evaluate the language-visual result. A higher CLIP score indicates a stronger correlation between the image and the description. The visual information is embedded to visual CLIP\cite{Radford} embedding $v$, and linguistic information is embedded to a textual CLIP embedding $c$, the relevance between the image and the language can be calculated as:

\begin{equation}
\text{CLIP}(c,v) = w \cdot \max(\cos(c,v), 0)
\end{equation}

By conducting these tests, we can evaluate the model’s capability to generate accurate visual representations from linguistic inputs and correctly label visual inputs with linguistic terms, thereby demonstrating its understanding of the concepts at various hierarchical levels.

\subsection{Language Understanding (Language-to-Vision)}

In the language understanding test, only the language information is provided to the model. We expect the model to be able to ``draw" an image based on the given concepts. Each of the three levels of hierarchical concepts is used as input to the model, with the image paired with the subordinate level concepts treated as the missing modality. The linguistic modality information is encoded by the encoder, and the decoder samples from the joint distribution to generate the missing visual modality.

Then the cross-generated image result is labeled by the pre-trained classifier, and the labels are compared with the model's concept input. Additionally, the generated image and input concepts are embedded to compute the CLIP score. As shown in Table 2, the pre-trained classifier's accuracy and the CLIP score for both levels of concept cross-generation results are similar to the ground truth, which is the accuracy and CLIP score of the dataset. This demonstrates that the model is capable of learning and understanding both concrete words and higher-order concepts. The generated results are shown in Figure 3.

\begin{table}
\caption{Result for the Language Understanding (Language-to-Vision) Test, the missing visual modality is generated based on the language modality. The generated image is then classified by the pre-trained classifier, and the accuracy is computed by comparing the classifier's predictions with the true labels. The ground truth is the accuracy of the classifier on the original dataset and the CLIP score of the original dataset.}\label{tab1}
\begin{tabular}{|>{\centering\arraybackslash}m{4cm}|>{\centering\arraybackslash}m{4cm}|>{\centering\arraybackslash}m{4cm}|}
\hline
 &  Classifier Accuracy & Clip Score\\
\hline
Subordinate Level & 90.29\% & 0.2790 \\
Basic Level & 99.34\% & 0.2624 \\
Subordinate Level Ground Truth  &87.45\% & 0.2756\\
Basic Level Ground Truth  &95.31\% & 0.2599\\

\hline
\end{tabular}
\end{table}
\begin{figure}
\includegraphics[width=0.8\textwidth]{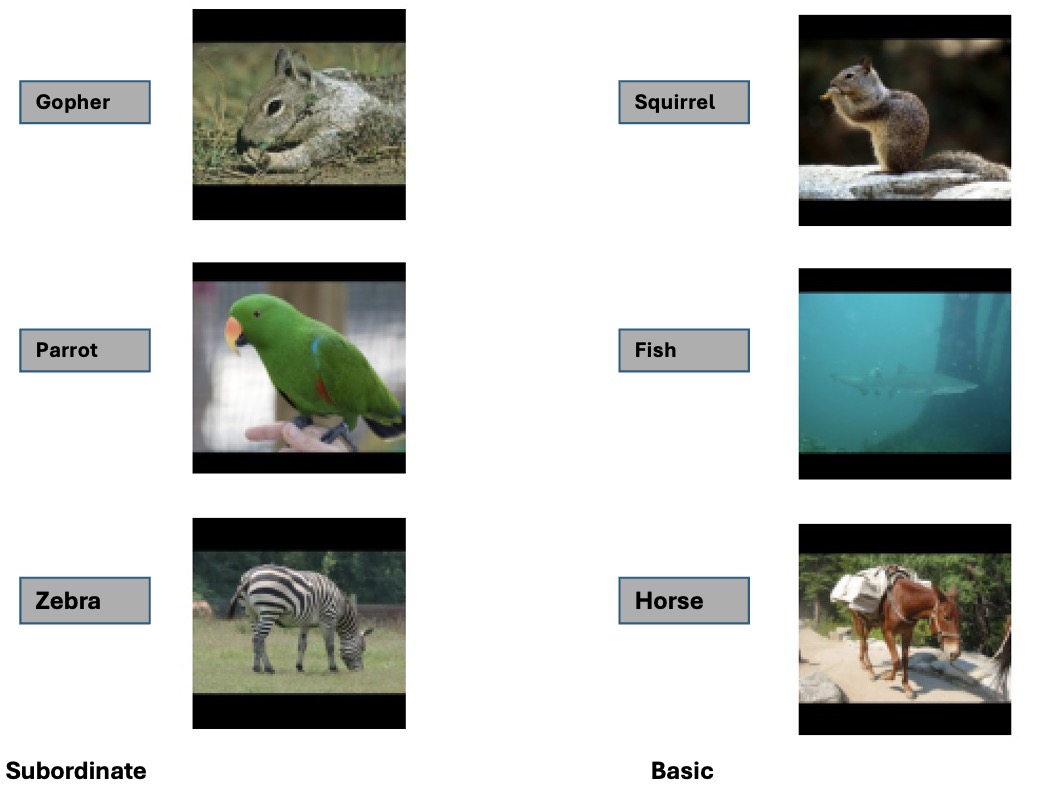}
\caption{The results for the Language Understanding (Language-to-Vision). The column on the left displays the images generated from the subordinate level concepts, while the column on the right shows the images generated from the basic level concepts.} \label{fig1}
\end{figure}
\subsection{Language Naming (Vision-to-Language)}

For the language naming test, only the visual information is provided. We expect the model to be able to ``name" the image. In this test, the image representing the visual information is the sole input to the model, and the model is expected to generate the corresponding linguistic modality information.
\begin{table}
\caption{Result for the Language Naming Test (Vision-to-Language), the missing language modality is generated based on the visual modality. The generated labels are compared with the true labels of the image to compute the accuracy of the label generation.}\label{tab1}
\begin{tabular}{|>{\centering\arraybackslash}m{4cm}|>{\centering\arraybackslash}m{4cm}|>{\centering\arraybackslash}m{4cm}|}
\hline
 &  Accuracy & Clip Score\\
\hline
Subordinate Level &81.25\%& 0.2717\\
Basic Level &92.91\%  &0.2558 \\
Subordinate Level Ground Truth  & 100\%  & 0.2756 \\
Basic Level Ground Truth  &100\%   & 0.2599\\
\hline
\end{tabular}
\end{table}
Similar to the language understanding experiment, the pre-trained classifier predicts the label of the model's input image, and the prediction is compared to the cross-generated concepts. Additionally, the CLIP score is computed by embedding the image and the generated concepts. As shown in Table 3, the results of the image-to-language experiment indicate that both the pre-trained classifier and the CLIP score show similar results for the image-generated label data and the original dataset. The generated results are shown in Figure 4.

\begin{figure}
\includegraphics[width=\textwidth]{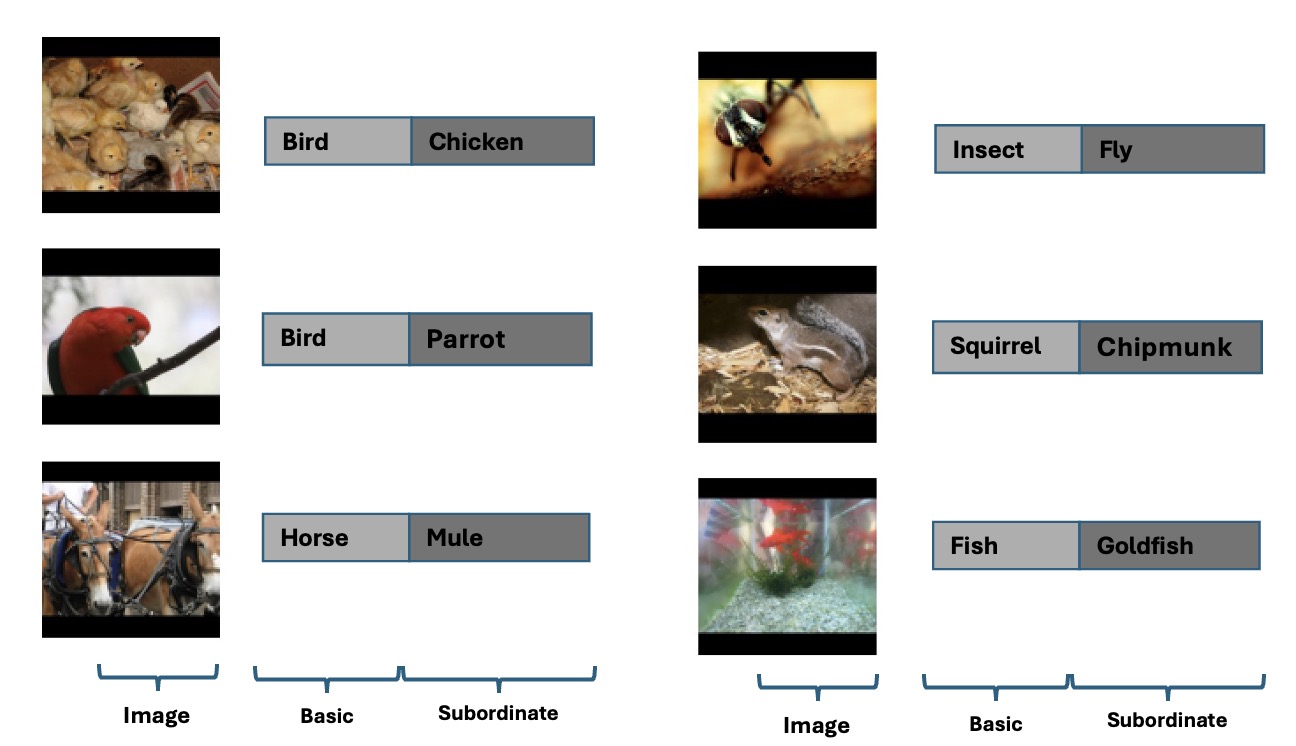}
\caption{The result for Language Naming (Vision-to-Language). The language modality is treated as the missing modality. The subordinate level labels and the basic level labels generated from the image.} \label{fig1}
\end{figure}
In the language-to-vision experiment, the classifier accuracy for the generated images is 90.29\% at the subordinate level, and 99.34\% at the basic level, both of which are higher than the classifier accuracy on the original dataset. This shows that the model is effective in generating images based on both levels of concepts. Additionally, both CLIP scores are similar to those of the original dataset, indicating a good correlation between the generated images and the concepts at both levels.

In the vision-to-language experiment, although the label generation accuracy for the subordinate level is relatively low at 81.25\%, the accuracy for basic level label generation is high at 92.91\%. Additionally, the CLIP scores for the generated labels at both levels and the images are similar to the scores for the original data, indicating a strong correlation between the generated labels and the images. This demonstrates the model's effectiveness in generating language from images.

Overall, the results of both experiments show that the model demonstrates good performance in learning and understanding both subordinate level and basic level concepts. Compared to pretrained vision-language models like CLIP, our model is trained on a much smaller dataset and uses only labels instead of sentence descriptions for training. 

\subsection{Ablation Studies}
In this section, we conduct further experiments to test the model's ability to learn with additional concepts. The original dataset is extended in two dimensions. As the experiment mentioned before, we performed the Language-to-Vision and Vision-to-Language tests with the two new datasets. The reconstruction results are evaluated with the CLIP score.

In the first experiment, two more subordinate level concepts are added to each basic-level category. In total, ten subordinate level concepts are added to the dataset: ``Grasshopper" and ``Ladybug" to ``Insects"; ``White Stork" and ``Ostrich" to ``Birds"; ``Guinea Pig" and ``Hamster" to ``Squirrel"; ``Lion fish" and ``Stingray" to ``Fish". Table 4 shows the CLIP score for the first dataset, which is comparable to the CLIP score for the original dataset. This indicates that the model is capable of learning and understanding concepts effectively, even with a larger dataset containing more subordinate level concepts. 
\begin{table}
\caption{The CLIP score for the first dataset which has 5 basic level concepts, each basic level concept contains 5 subordinate level concepts.}\label{tab1}
\begin{tabular}{|>{\centering\arraybackslash}m{4cm}|>{\centering\arraybackslash}m{4cm}|>{\centering\arraybackslash}m{4cm}|}
\hline
 &  Language-to-Vision & Vision-to-Language\\
\hline
Subordinate Level &0.2780& 0.2747\\
Basic Level &0.2569  &0.2504 \\
Subordinate Level Ground Truth  & 0.2776  & 0.2776 \\
Basic Level Ground Truth  & 0.2532   & 0.2532\\
\hline
\end{tabular}
\end{table}

In the second experiment, two additional basic-level concepts, ``Cat" and ``Dog", are added to the original dataset. Like the original dataset, each basic-level concept includes three subordinate level concepts. For ``Cat", the subordinate level concepts `Tiger cat", ``Egyptian cat" and ``Persian cat" are added, while for ``Dog", the subordinate level concepts `English Foxhound", ``Border Collie" and ``Golden Retriever" are included. Table 5 shows the CLIP score for the first dataset. The results further demonstrate that the model can effectively learn and understand concepts when trained on a larger dataset with more basic-level concepts.

\begin{table}
\caption{The CLIP score for the second dataset which has 7 basic level concepts, each basic level concept contains 3 subordinate level concepts.}\label{tab1}
\begin{tabular}{|>{\centering\arraybackslash}m{4cm}|>{\centering\arraybackslash}m{4cm}|>{\centering\arraybackslash}m{4cm}|}
\hline
 &  Language-to-Vision & Vision-to-Language\\
\hline
Subordinate Level &0.2734 & 0.2722\\
Basic Level &0.2539  &0.2519 \\
Subordinate Level Ground Truth  & 0.2786  & 0.2786 \\
Basic Level Ground Truth  & 0.2559   & 0.2559\\
\hline
\end{tabular}
\end{table}

Overall, the results from the two additional experiments show that the proposed model is able to learn and understand concepts with a larger dataset as the number of concepts increases. This suggests it could be an effective approach for incorporating more concrete and abstract concepts in the acquisition of abstract words.
\section{Conclusion}

In this research, we propose a multimodal generative model for learning both concrete and abstract concepts by integrating visual and linguistic information. Our findings show that the proposed MMVAE-based model effectively learns and generates higher-order abstract concepts by grounding them with the knowledge of concrete concepts. Through cross-generation testing, we demonstrate that the model not only accurately reconstructs visual and linguistic inputs but also effectively bridges the gap between concrete and abstract concept understanding. Unlike pre-trained vision-language models such as CLIP, which map images to texts and vice versa, our model learns a joint distribution to represent the information from both visual and language modalities. The learning process closely resembles human cognitive development. In this way, our work proposes a way to include more concrete and abstract concepts for concept learning tasks with a joint distribution that represents all multimodal information.

Due to computational power and time limitations, additional computational resources and time are needed to include more concepts for more abstract concept learning. At this stage, our model in this work is limited to one superordinate level concept. In future work, we plan to extend the proposed model to learn more abstract concepts by including developmental linguistics datasets and age-of-acquisition lexical data like \cite{Morrison}. By incorporating developmental data \cite{Morrison}, the model can prioritize learning earlier-acquired concepts before progressing to later-acquired ones. This approach will enable the model to understand more abstract and later-acquired concepts to be built upon the foundation of earlier-acquired concepts. At the same time, in this work, the reconstruction of visual information is based on a neighborhood search in the feature space learned solely from the visual modality. A potential improvement to the proposed method could involve using an embedding representation of a multimodal model.

\begin{credits}
\subsubsection{\ackname} This paper was partially funded by the ERC Advanced project eTALK (funded by UKRI).
\end{credits}

%
%
%
%

\end{document}